\documentclass[runningheads]{llncs}

\usepackage{eccv}

\usepackage{eccvabbrv}

\usepackage{graphicx}
\usepackage{booktabs}
\usepackage{multirow} 
\usepackage{makecell}
\usepackage{colortbl}  
\usepackage{array}   
\usepackage{marvosym}
\usepackage[accsupp]{axessibility}  
\usepackage{wrapfig}

\usepackage{hyperref}

\usepackage{orcidlink}

\begin{document}

\title{DA-BEV: Unsupervised Domain Adaptation for Bird’s Eye View Perception} 

\titlerunning{DA-BEV}

\author{
Kai Jiang \inst{1,}$^\dag$ \orcidlink{0000-0001-9921-2043} \and
Jiaxing Huang \inst{2,}$^\dag$ \orcidlink{0000-0002-8681-0471} \and
Weiying Xie\inst{1} \orcidlink{0000-0001-8310-024X} \and
Jie Lei\inst{4} \orcidlink{0000-0003-0851-6565} \and
Yunsong Li\inst{1} \and
Ling Shao\inst{3} \orcidlink{0000-0002-8264-6117} \and
Shijian Lu\inst{2,}$^*$ \orcidlink{0000-0002-6766-2506}
}

\authorrunning{K.~Jiang et al.}

\institute{State Key Laboratory of Integrated Services Networks, Xidian University, Xi’an 710071, China. 
\email{kjiang\_19@stu.xidian.edu.cn}\\
\and
S-lab, School of Computer Science and Engineering, Nanyang Technological University.
\email{\{Jiaxing.Huang, Shijian.Lu\}@ntu.edu.sg}\\
\and
UCAS-Terminus AI Lab, University of Chinese Academy of Sciences, China. \\
\and
School of Electrical and Data Engineering at the University of Technology Sydney.
}

\renewcommand{\thefootnote}{}

\footnotetext[1]{$^\dag$ These authors contributed equally to this work.} 
\footnotetext[2]{$^*$ Corresponding author.}

\maketitle

\begin{abstract}
Camera-only Bird's Eye View (BEV) has demonstrated great potential in environment perception in a 3D space. However, most existing studies were conducted under a supervised setup which cannot scale well while handling various new data. Unsupervised domain adaptive BEV, which effective learning from various unlabelled target data, is far under-explored. In this work, we design DA-BEV, the first domain adaptive camera-only BEV framework that addresses domain adaptive BEV challenges by exploiting the complementary nature of image-view features and BEV features. DA-BEV introduces the idea of query into the domain adaptation framework to derive useful information from image-view and BEV features. It consists of two query-based designs, namely, query-based adversarial learning (QAL) and query-based self-training (QST), which exploits image-view features or BEV features to regularize the adaptation of the other. Extensive experiments show that DA-BEV achieves superior domain adaptive BEV perception performance consistently across multiple datasets and tasks such as 3D object detection and 3D scene segmentation.
\end{abstract}

\section{Introduction}

\begin{wrapfigure}[28]{r}{0.45\textwidth}
    \centering
    \includegraphics[width=0.45\textwidth]{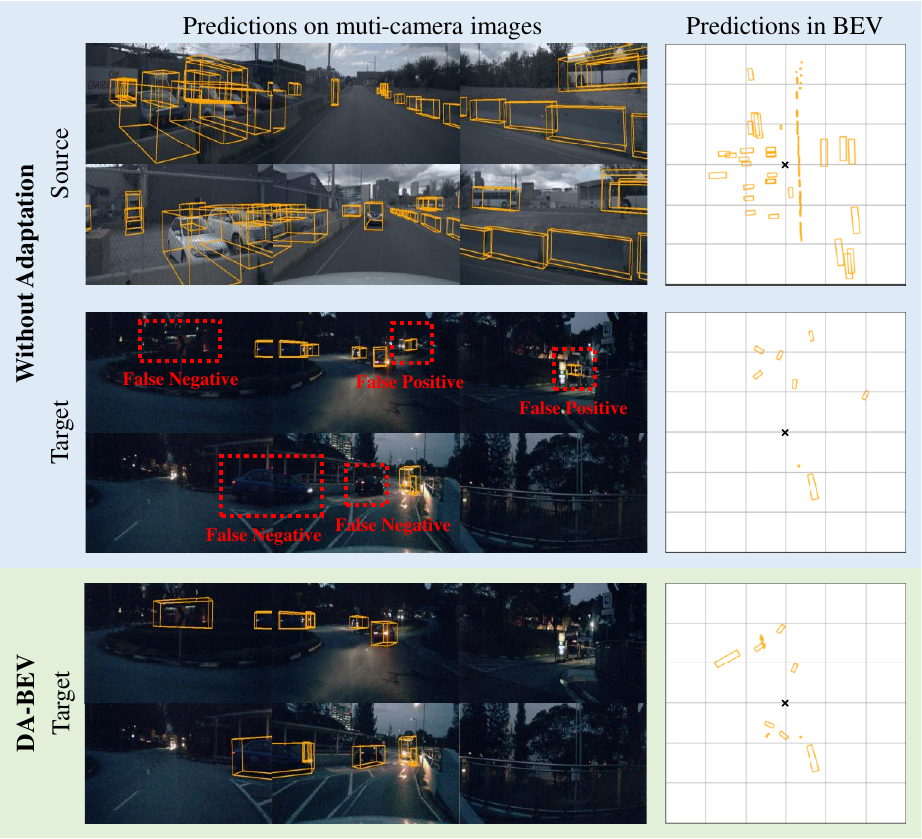}
    \caption{
    Domain adaptive bird’s eye view perception (DA-BEV). The first two rows show the detection by a source-only model (trained with source data with no adaptation) over a source scene and a target scene, respectively. The yellow 3D boxes indicate correct detection while the red dotted boxes highlight false-positive and false-negative detection. The third row shows the detection by our DA-BEV on the same target scene. The two columns visualize the 3D predictions in multi-camera view and bird’s eye view, respectively.}\label{fig:figure1}
\end{wrapfigure}

3D visual perception~\cite{ma2022vision,li2022delving} aims at sensing and understanding surrounding environments in 3D space, which plays an important role in various applications such as mobile robotics~\cite{antonello2017fast, gutmann20083d}, autonomous driving~\cite{wang2023multi,wangYingjie2023multi}, virtual reality~\cite{schuemie2001research,zhao2009survey}, etc.
Despite the remarkable progress of monocular and LiDAR-based 3D perception~\cite{xu2021rpvnet,guo2021liga,bai2022transfusion, li2022deepfusion, jiao2023msmdfusion, zhu2021cylindrical, yin2021center,yoo20203d,vora2020pointpainting,chen2022autoalignv2,chen2022autoalign,wang2021pointaugmenting}, camera-only 3D perception in Bird's Eye View (BEV)~\cite{li2022bevformer, liu2022petr, jiang2023polarformer, liu2023petrv2, yang2023bevformer, Xiong2023CAPE} has drawn increasing attention in recent years thanks to its advantages in comprehensive 3D understanding, rich semantic information, high computational efficiency, and low deployment cost.
On the other hand, camera-only BEV models~\cite{wang2023towards} trained over a source domain usually experience clear performance degradation when applied on a target domain due to clear cross-domain discrepancy as illustrated in Fig.~\ref{fig:figure1}. 
Unsupervised Domain adaptation (UDA)~\cite{wilson2020survey} aims to address the cross-domain discrepancy by transferring a deep network model trained over a labelled source domain towards an unlabelled target domain. 
It has been extensively explored for various 2D computer vision tasks~\cite{zhang2021survey,oza2023unsupervised,Gao_2023_ICCV,Zhu_2023_CVPR} that perceive environments in a 2D space.
However, how to mitigate the cross-domain discrepancy for camera-only BEV perception is much more challenging but largely under explored.

\begin{wrapfigure}[18]{r}{0.45\textwidth}
    \centering
    \includegraphics[width=0.45\textwidth]{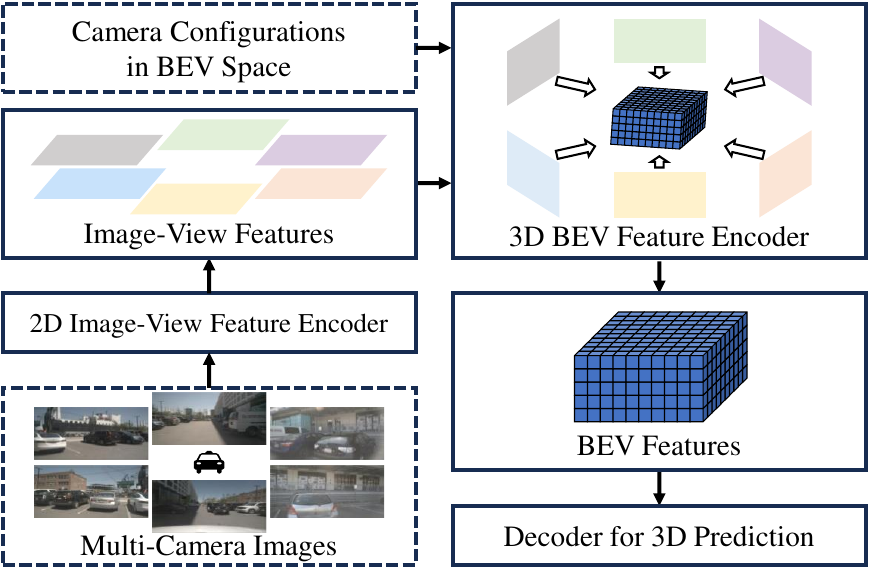}
    \caption{
The architecture of camera-only BEV models. 
Boxes in dash lines denote model inputs including multi-camera images and camera configurations. Boxes in solid lines stand for encoding/decoding processes and intermediate representations.
}   \label{fig:intro}
\end{wrapfigure}

We explore unsupervised domain adaptation for camera-only BEV perception, and we start from a simple observation: Camera-only BEV perception generally works in two stages by first encoding multi-camera images into image-view features separately and then fusing the image-view features with the corresponding camera configurations into BEV features.
As shown in Fig.~\ref{fig:intro}, BEV features can capture global 3D information over the BEV space but they usually suffer from more severe inter-domain discrepancy due to explicit changes of camera configurations beyond the change of image styles and appearance within the image-view features.
As a comparison, the image-view features are encoded from images only and would not be explicitly affected by camera configurations , which suffer from less inter-domain discrepancy but capture largely local information within each single image and lack global 3D information.
Hence, one effective way of bridging the inter-domain discrepancy of camera-only BEV perception is by exploiting the complementary nature of image-view features and BEV features.

To this end, we design DA-BEV, the first domain adaptive camera-only BEV perception framework that addresses BEV domain adaptation challenges by exploiting the complementary nature of image-view features and BEV features. We introduce learnable queries to facilitate the interplay between image-view and BEV features while adapting them across domains. Specifically, the global 3D information from BEV features helps adapt image-view features while the 2D information in image-view features with less domain variation helps adapt BEV features. Following this philosophy, we design two query-based domain adaptation techniques including query-based adversarial learning (QAL) and query-based self-training (QST): QAL/QST exploits the useful information queried from image-view features or BEV features to regularize the ``adversarial learning''/``self-training'' of the another.

The proposed DA-BEV has three desirable features: 1) It introduces a novel query-based domain adaptation strategy that exploits the complementary nature of image-view features and BEV features for unsupervised BEV perception adaptation. The query-based domain adaptation strategy is well aligned with the architecture of camera-only BEV perception models; 2) It designs query-based adversarial learning and query-based self-training that achieve UDA by aligning inter-domain features and maximizing the target-domain likelihood, respectively. The two designs complement each other which jointly achieve robust unsupervised BEV perception adaptation effectively; 3) It is generic and can work for different camera-only BEV perception tasks such as 3D object detection and 3D scene segmentation.

The contributions of this work can be summarized in three major aspects. \textit{First}, we explore unsupervised camera-only BEV perception adaptation and propose a query-based technique that mitigates the inter-domain discrepancy by exploiting the complementary nature of image-view features and BEV features. To the best of our knowledge, this is the first work that exploits the specific BEV network architecture for unsupervised BEV perception adaptation. \textit{Second}, we design DA-BEV that introduces query-based adversarial learning and query-based self-training that jointly tackle domain adaptive BEV perception effectively. \textit{Third}, extensive experiments show that DA-BEV achieves superior BEV perception adaptation consistently across different datasets and tasks such as 3D object detection and 3D scene segmentation.

\section{Related Work}
\subsection{Camera-only BEV Perception}
Camera-only Bird's Eye View (BEV) perception~\cite{li2022delving, ma2022vision} has recently attracted increasing attention thanks to its rich semantic information, high computational efficiency, and low deployment cost.
It captures multi-view images of a scene and formulates unified BEV feature in ego-car coordinates for various 3D vision tasks such as 3D detection, map segmentation, etc.
According to the way of view transformation, most existing camera-only BEV methods can be grouped into four categories~\cite{ma2022vision}, namely, homograph-based, depth-based, multilayer perception (MLP)-based, and transformer-based.
Homograph-based methods~\cite{song2021stacked,hou2020multiview,garnett20193d,reiher2020sim2real,loukkal2021driving} directly transform the multi-view image features into BEV features with explicitly constructed homography matrices.
The recent are computationally efficient but exhibit limited performance in complex real-world scenarios. 
Depth-based methods~\cite{wang2022mv,zhang2022beverse, huang2022bevdet4d, li2023bevdepth,huang2021bevdet,sun2022putting,akan2022stretchbev,xie2022m,hu2021fiery,chen2020dsgn} estimate depth distribution and apply it to generate BEV features from multi-view 2D features. 
MLP methods~\cite{yang2021projecting,saha2021enabling,roddick2020predicting, pan2020cross,zou2022hft,li2022hdmapnet,hendy2020fishing} learn a transformation between multi-view features and BEV features, aiming to relieve the inherent inductive biases under a calibrated camera setup.
The recent transformer-based methods~\cite{wang2022detr3d, li2022bevformer, liu2022petr, jiang2023polarformer,liu2023petrv2,yang2023bevformer,wang2022focal,Xiong2023CAPE,xu2022cobevt,roh2022ora3d,shi2022srcn3d,bartoccioni2023lara,chen2022polar,chen2022graph,chen2022persformer,peng2023bevsegformer} design a set of BEV queries to retrieve the corresponding image features via cross-attention, ultimately constructing BEV features in a top-down manner. 
Differently, we explore unsupervised domain adaptation for camera-only BEV perception. 

\subsection{Domain Adaptation}
{Domain Adaptation} aims to adapt source-trained models towards various target domains.
Most existing work focuses on unsupervised domain adaptation (UDA) for 2D vision tasks, which minimizes the domain discrepancy by discrepancy minimization~\cite{long2015learning, vu2019advent}, adversarial training~\cite{gong2019dlow, vu2019advent, luo2021category}, or self-training~\cite{lee2013pseudo, zhang2019category, zhang2021prototypical}.
However, most existing 2D UDA methods do not work well in 3D vision tasks that aim to understand objects and scenes in three-dimensional space.
3D UDA has been explored recently. For example, STMono3D~\cite{li2022unsupervised} introduces self-training
for cross-domain monocular 3D object detection.
\cite{saleh2019domain, barrera2021cycle} and~\cite{acuna2021towards} exploit lidar and multimodal data in 3D UDA but they cannot work for camera-only BEV perception with extra data modality. 
Recently, $\text{M}^2\text{ATS}$~\cite{liu2022multi} studies domain adaptation for multi-view 3D detection but still relies on depth supervision from Lidar. 
Unsupervised domain adaptation has been largely neglected for the task of camera-only BEV perception

\section{Domain Adaptive BEV Perception}
\subsection{Preliminary}
\textbf{Task definition.} 
This work focuses on the task of unsupervised domain adaptation of camera-only BEV perception. It involves a labelled source domain $D_S=\{x^n_S, k^n_S, y^n_S\}_{n=1}^{N_S}$ and an unlabelled target domain $D^T=\{x^n_T, k^n_T\}_{n=1}^{N_T}$, where $x$ denotes a set of multi-camera images, $k$ stands for the camera configurations, and $y$ denotes the 3D object annotations in the BEV space as represented by $x$ and $k$. 
The goal is to learn a BEV perception model that performs well on the unlabelled target domain.
\textbf{A Revisit of BEV Models.}
Camera-only BEV model typically works in two stages as illustrated in Fig.~\ref{fig:intro}. It first encodes multi-camera images into image-view features and then fuses these image-view features with the corresponding camera configurations to produce BEV features for 3D perception.
Specifically, camera-only BEV model consists of a 2D image-view feature encoder $\mathcal{E}^{iv}$, a \textcolor{black}{3D BEV feature encoder} $\mathcal{E}^{bev}$, a BEV feature decoder $\mathcal{D}^{bev}$, and a 3D detection head $\mathcal{H}_{det}$.
Given a set of multi-camera images $x$ and their corresponding camera configurations $k$, camera-only BEV model first encodes $x$ into image-view features with the $\mathcal{E}^{iv}$ as follows:

\begin{equation}
    {f^{iv}} = \mathcal{E}^{iv}(x),
\end{equation}
The image features are then encoded with the corresponding camera configurations $k$ into BEV features with the $\mathcal{E}^{bev}$:
\begin{equation}
    {f^{bev}} = \mathcal{E}^{bev}\left(f^{iv}, k\right).
\end{equation}
The BEV feature decoder $\mathcal{D}^{bev}$ then decodes the BEV features $f^{bev}$ with a set of learnable BEV queries $q^{bev}$:
\begin{equation}
\label{PETRHead}
    \tilde{q}^{bev}=\mathcal{D}^{bev}(q^{bev}, f^{bev}).
\end{equation}
where the BEV queries ${q}^{bev}$ interact with the BEV features $f^{bev}$ and aggregate the 3D location information of objects in BEV,  leading to the decoded query features $\tilde{q}^{bev}$.  
Given the decoded query features $\tilde{q}^{bev}$, 3D detection head $\mathcal{H}_{det}$ predicts 3D object bounding boxes $p^{det}$ by:
\begin{equation}
\label{eq:detection_head}
    p^{det}=\mathcal{H}_{det}(\tilde{q}^{bev}).
\end{equation}
Under the setting of supervised learning, the camera-only BEV model can be formulated as follows:
\begin{equation}
    {L}^{bev} = \mathcal{L}_{det}(p^{det}, y).
\end{equation}
where $\mathcal{L}_{det}$ denotes a 3D detection loss function.


\subsection{{Framework of DA-BEV}}

\begin{figure}[tb]
\centering
\includegraphics[width=1.0\linewidth]{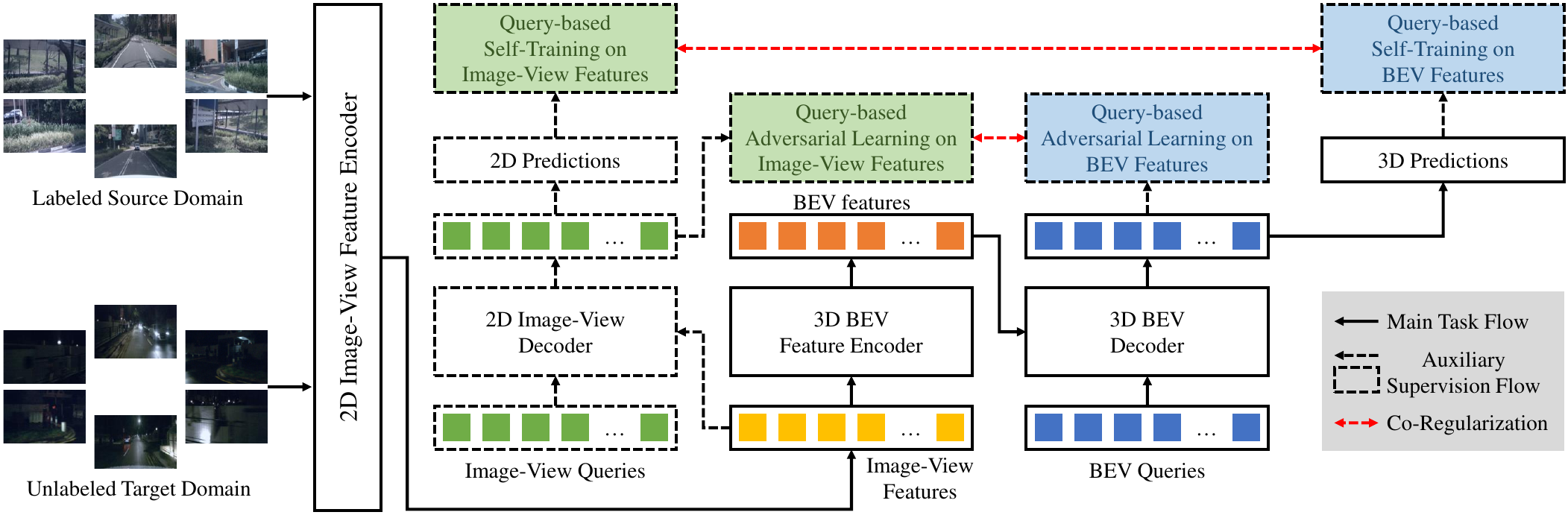}
\caption{
The overall framework of DA-BEV, which exploits the complementary nature of image-view and BEV features for unsupervised BEV perception adaptation.
To this end, DA-BEV first introduces an additional 2D Image-View Decoder into the BEV perception model to capture image-view features with local 2D information, which complement BEV features that capture rich global 3D information.
The training of DA-BEV comprises two designs including query-based adversarial learning (QAL) and query-based self-training (QST). 
The former exploits the complementary information from image-view or BEV features to regularize the adversarial learning of the another, while the latter exploits the complementary information from both image-view and BEV features to regularize their self-training.
Note, all the auxiliary supervision flows are utilized during network training but discarded after adaptation, which introduces slight computation overhead during model training but has little effect on testing. 
}
\label{fig:schematic}
\end{figure}

Our method tackles BEV domain adaptation challenges by exploiting the complementary nature of image-view features and BEV featuresas illustrated in Fig.~\ref{fig:schematic}. 
Specifically, we design novel query-based domain adaptation that introduces learnable queries to enable interaction between the image-view and BEV features and as well as their synergetic adaptation.
Intuitively, the global 3D information from BEV features helps adapt image-view features while the local 2D information in image-view features with less domain variation helps adapt BEV features.
Based on this philosophy, we design two query-based domain adaptation techniques, namely, query-based adversarial learning (QAL) and query-based self-training (QST). 
QAL/QST exploits the useful information queried from image-view features or BEV features to regularize the ``adversarial learning''/``self-training'' of the another as illustrated in Fig.~\ref{fig:schematic}.
To capture 2D information in the image-view features with less inter-domain, we introduce an image-view feature decoder $\mathcal{D}^{iv}$ with a set of learnable image-view queries $q^{iv}$. 
The interaction between $q^{iv}$ $f^{iv}$ thus produces the image-view query features $\tilde{q}^{iv}$ as follows:
\begin{equation}
\label{eq:en_transoformer_de}
    \tilde{q}^{iv} = \mathcal{D}^{iv}(q^{iv},{f}^{iv}).
\end{equation}
The query feature $\tilde{q}^{iv}$ is then fed into a multi-label classification head $\mathcal{H}_{cls}$ to predict the probability of each object category, where $\mathcal{H}_{cls}$ is trained by a multi-label classification loss function ${L}_{cls}$ as follows:
\begin{equation}
\label{eq:image_view_prediction}
    p^{cls}=\mathcal{H}_{cls}(\tilde{q}^{iv}), \ \ \ {L}^{iv} = \mathcal{L}_{cls}(p^{cls}, y^{iv}),
\end{equation}
where $y^{iv}$ denotes the image-view multi-label classification annotations. Note $\mathcal{L}_{cls}$ is computed for both source and target data.
For source data $x_S, k_S, y_S$, we generate the image-view multi-label classification annotation $y^{iv}_{S}$ by merging the object category labels in $y_S$. For target data $x_T, k_T$, we generate $y^{iv}_{T}$ based on the model predictions as defined in Eq.~\ref{eq:QST}.  

Hence, the decoded image-view query features $\tilde{q}^{iv}$ captures rich 2D semantic and location information from the image-view features.
They contain much less inter-domain discrepancy than the BEV features and can therefore benefit BEV feature adaptation greatly.
In addition, the multi-label classification training objective of image-view query $q^{iv}$ encourages capturing more balanced information among different categories, which facilitates the adaptation of BEV feature which are trained with one-hot object category annotation and thus often suffer from clear class imbalance issues~\cite{chen2023revisiting}.

To capture the global 3D information in BEV features, we directly employ the off-the-shelf BEV queries $q^{bev}$ in Eq.~\ref{PETRHead}, which interact with BEV features to generate the decoded BEV query features $\tilde{q}^{bev}$. 
As BEV features are encoded with camera configurations and $q^{bev}$ is trained with 3D object annotations, the decoded BEV query feature $\tilde{q}^{bev}$ captures rich global 3D information including the object position in 3D BEV space. 
They can help the adaptation of image-view features that capture little global 3D information in BEV space.

\subsection{Query-based Adversarial Learning}
\label{section:Query-based Adversarial Learning}
Our proposed QAL exploits the useful information queried from image-view or BEV features to regularize the adversarial learning of the another. 
As illustrated in Fig.~\ref{fig:schematic}, QAL employs two domain classifiers to measure the inter-domain distances of the image-view query features and the BEV query features respectively, and utilizes the measured inter-domain distance for mutual regularization.

Hence, QAL simultaneously mitigates the inter-domain discrepancies of the local 2D information in image-view features and the global 3D information in BEV features, both being critical to locate and recognize objects and backgrounds in a 3D space.
Besides, the adversarial learning of 2D image-view query features involves little 3D information, where the BEV query features can regularize it effectively by providing rich global 3D information, i.e.,  
enhancing 2D image-view adversarial learning for the samples with significant 3D BEV inter-domain discrepancies while diminishing 2D image-view adversarial learning for the samples with minor 3D BEV inter-domain discrepancies. 
Similarly, the adversarial learning of 3D BEV query features can be regularized by the image-view query features capturing rich 2D semantic and location information.
Specifically, we employ domain classifiers $\mathcal{C}^{iv}$ and $\mathcal{C}^{bev}$ to measure the inter-domain distance of 2D image-view and 3D BEV features as follows:
\begin{equation}
    \begin{aligned}
    &\lambda^{iv}=\text{exp}\left(-\text{log}\mathcal{C}^{iv}(\tilde{q}^{iv}_s)-\text{log}\left(1-\mathcal{C}^{iv}(\tilde{q}^{iv}_t)\right)\right),\\ 
    &\lambda^{bev}=\text{exp}\left(-\text{log}\mathcal{C}^{bev}(\tilde{q}^{bev}_s)-\text{log}\left(1-\mathcal{C}^{bev}(\tilde{q}^{bev}_t)\right)\right),
    \end{aligned}
\end{equation}
where $\tilde{q}^{iv}_s$ and $\tilde{q}^{iv}_t$ denote the decoded image-view query features in Eq.~(\ref{eq:en_transoformer_de}) that interacted with the source and target image-view features, respectively.
$\tilde{q}^{bev}_s$ and $\tilde{q}^{bev}_t$ denote the decoded BEV query features in Eq.~(\ref{PETRHead}) that interacted with source and target domain BEV features, respectively.


The mutual regularization of QAL between image-view and BEV features can be formulated as follows,
\begin{equation}
{L}_{qal}={\lambda}^{iv}\cdot \mathcal{L}_{al}(\tilde{q}^{iv}_s, \tilde{q}^{iv}_t))+ {\lambda}^{bev}\cdot \mathcal{L}_{al}(\tilde{q}^{bev}_s,\tilde{q}^{bev}_t),
\end{equation}
where $\mathcal{L}_{al}(\cdot)$ is a widely adopted adversarial learning loss function for cross-domain alginment~\cite{ganin2016domain}.

\subsection{Query-based Self-training}
\label{section:Query-based Self-training}

Our proposed QST exploits the useful information queried from both image-view and BEV features to regularize their self-training, as shown in Fig.~\ref{fig:schematic}.
Intuitively, the decoded image-view query features capture rich 2D semantic and location information that has less inter-domain discrepancy, while the decoded BEV query features capture rich global 3D information in the BEV space. 
Hence, the two types of features complement each other and together regularize self-training effectively with comprehensive 2D and 3D information.
Specifically, QST first leverages the predictions from image-view or BEV features to denoise the predictions from the other:
\begin{equation}    \hat{p}^{cls}=\text{max}\left(p^{det}\right)\cdot p^{cls},\ \ \ 
    \hat{p}^{det}=p^{det} \cdot p^{cls},
\end{equation}
where $p^{cls}/\hat{p}^{cls}$ and $p^{det}/\hat{p}^{det}$ denotes image-view and BEV predictions before/after cross-view denoising. 
The operation $\text{max}(\cdot)$ stands for selecting the maximum probability in BEV predictions for each category.
Please note we conduct the above cross-view denoising operations for each camera perspective, respectively.
Then, QST acquires global class distribution by accumulating the denoised predictions that capture comprehensive 2D and
3D information, and further exploit it to facilitate pseudo label generation.
Specifically, we model the global class probability distribution as Gaussian distributions $\mathcal{N}(\mu_c,\sigma_c^2)$, where $c$ denotes the category index.
We estimate $\hat{\mu}_c$ and $\hat{\sigma}^2_c$ by maintaining a first-in-first-out queue of historical predictions:
\begin{equation}
\label{eq:batch_estimate_mu}
    \hat{\mu}_c=\frac{1}{B}\sum_{j=1}^{B}\hat{p}^{det}_c(j), \ \ 
    \hat{\sigma}^2_c=\frac{1}{B-1}\sum_{j=1}^{B}\left(\hat{p}^{det}_c(j)-\hat{\mu}_c\right)^2,
\end{equation}
where $B$ denotes the queue length and $\hat{p}^{det}_c(j)$ stands for the $j$-th historical prediction in the queue. 
Note we use $\hat{p}^{det}$ only as it already includes the 2D information in $\hat{p}^{cls}$ after cross-denoising.

We accumulate $\hat{\mu}_c$ and $\hat{\sigma}^2_c$ along the training process for more stable estimation of $\mathcal{N}({\mu}_c,{\sigma}_c^2)$:
\begin{equation}
\label{eq:ema_estimate_mu}
\mu_c= \gamma \cdot \mu_c + (1-\gamma)\cdot \hat{\mu}_c,\ \ \sigma^2_c=\gamma \cdot \sigma^2_c + (1-\gamma)\cdot \hat{\sigma}^2_c,
\end{equation}
where $\gamma$ denotes the Exponential Moving Average parameter for smooth update.

According to the estimated global class probability distribution, we select the top $\upsilon$ percent predictions as the pseudo labels for each category. The pseudo label threshold $\tau_c$ can be determined by solving the following equation:
\begin{equation}
\label{eq:pseudolabelthresholdcal}
    \frac{\Phi(1|\mu_c,\sigma_c)-\Phi(\tau_c|\mu_c,\sigma_c)}{\Phi(1|\mu_c,\sigma_c)-\Phi(0|\mu_c,\sigma_c)}=\upsilon,
\end{equation}
where $\Phi(\cdot)$ refers to the cumulative density function of $\mathcal{N}(\mu_c,\sigma_c^2)$. Note we apply thresholds $\{\tau_c|c=1,...,C\}$ to both image-view and BEV predictions $\hat{p}^{cls}$ and $\hat{p}^{det}$ to acquire image-view and BEV pseudo labels $\hat{y}^{cls}$ and $\hat{y}^{det}$.

Our pseudo label generation method thus has three desirable features: 1) the thresholds are update-to-date as they are dynamically determined along the training process according to the complementary 2D and 3D information captured by image-view and BEV features; 2) it mitigates class imbalance issue by selecting the same percentage of pseudo labels for each class; 
3) it works online and does not need an additional inference round as in~\cite{zou2018unsupervised,jamal2020rethinking}.



The training loss of QST can thus be formulated by:
\begin{equation}
\label{eq:QST}
    {L}_{qst}=\mathcal{L}_{cls}(p^{cls},\hat{y}^{iv})+\mathcal{L}_{det}(p^{det},\hat{y}^{bev}).
\end{equation}

\subsection{Overall Objective}
In summary, the overall training objective of the proposed DA-BEV can be formulated as follows:
\begin{equation}
\mathop{\text{min}}_{\mathcal{E},{\mathcal{D}},\mathcal{H}}\mathop{\text{max}}_{\mathcal{C}} {L}^{iv} +{L}^{bev} +  {L}_{qst}+{L}_{qal},
\end{equation}
where $\mathcal{E}=\{\mathcal{E}^{2d},\mathcal{E}^{3d}\}$, $\mathcal{D}=\{\mathcal{D}^{iv}, \mathcal{D}^{bev}\}$, $\mathcal{H}=\{\mathcal{H}_{cls},\mathcal{H}_{det}\}$, $\mathcal{C}=\{\mathcal{C}^{iv}, \mathcal{C}^{bev}\}$.

\section{Experiments}

\subsection{Datasets}
We benchmark our DA-BEV extensively over various BEV adaptation scenarios including adaptation across different illumination conditions, different weather conditions, different cities, and different camera systems. 
The four BEV adaptation scenarios involve two 3D perception datasets including nuScenes~\cite{caesar2020nuscenes} and Lyft~\cite{houston2021one}.

\textbf{nuScences~\cite{caesar2020nuscenes}} is a large-scale autonomous driving 3D dataset, which comprises 6-camera images captured from 1000 scenes, including the scenes with daytime/nighttime and ``clear weather''/``rainy weaether'' conditions, and from Singapore/Boston cities.
For 3D object detection, it includes 1.4M annotated 3D bounding boxes with 10 categories.
For 3D scene segmentation~\cite{zhou2022cross, liu2023petrv2}, it contains 200K frames with fine-grained segmentation annotations in 3D space.
\textbf{Lyft~\cite{houston2021one}} is a self-driving dataset for 3D perception, which has a diverse set of cameras with distinct intrinsic and extrinsic parameters.
It comprises 366 scenes with 1.3M 3D object detection annotations, which are captured from California, and under a different camera system as compared with nuScences.
With nuScences~\cite{caesar2020nuscenes} and Lyft~\cite{houston2021one}, the six BEV adaptation scenarios are formulated.  
We formulate ``nuScences Daytime $\rightarrow$ nuScences Nighttime'' for cross-illumination adaptation, ``nuScences Clear Weather $\rightarrow$ nuScences Rainy Weather'' for cross-weather adaptation,
and ``nuScences Singapore $\rightarrow$ nuScences Boston'' for cross-city adaptation. 
Moreover, ``Lyft $\rightarrow$ nuScences'', ``Lyft $\rightarrow$ nuScences Nighttime'', and ``Lyft $\rightarrow$ nuScences Rainy Weather'' are formulated for adaptation across different camera configurations.

\begin{table}[tb]
\caption{
Unsupervised domain adaptation of camera-only BEV perception across different illumination conditions, weather conditions, cities. 
}
\renewcommand{\arraystretch}{0.1}
\tiny
\centering
\resizebox{\linewidth}{!}{
\begin{tabular}{c|cccccccccc|cc}
\toprule
\multicolumn{13}{c}{{\textbf{Daytime $\rightarrow$ Nighttime}}} \\
\midrule
\multirow{3}*{Methods} & \multicolumn{10}{c|}{Average Precision$\uparrow$} &\multirow{3}*{ mAP$\uparrow$} & \multirow{3}*{NDS$\uparrow$}\\
\cmidrule{2-11}
&Car&Truck&{Cons. Vehicle}&Bus&Trailer&Barrier&Motorcycle&Bicycle&Pedestrian&{Cone}&&  \\

\midrule
PETR~\cite{liu2022petr} (Baseline)&40.79   &28.88   &0.41    &29.71   &0.00    &9.15  &22.00   &0.71    &24.66    &2.84    &15.92  &22.44 \\
SFA~\cite{wang2021exploring} &40.70 	&27.05 	&0.00 	&30.48 	&0.00 	&11.04 	&21.52 	&0.42 	&24.21 	&4.69 	&16.01 &23.95 
 \\
MTTrans~\cite{yu2022mttrans} &41.48 	&34.39 	&0.01 	&30.49 	&0.00 	&12.48 	&21.40 	&0.69 	&24.94 	&4.16 	&17.00 &24.48 
 \\
STM3D~\cite{li2022unsupervised} &41.77 	&33.19 	&0.27 	&29.22 	&0.00 	&13.28 	&21.94 	&0.28 	&24.21 	&4.36 	&16.85 	&24.41  \\
DA-BEV (Ours) &45.07 	&41.96 	&1.91 	&38.12 	&0.00 	&16.66 	&25.74 	&0.54 	&28.05 	&4.65 	&20.27&26.98\\
\end{tabular}
}
\resizebox{\linewidth}{!}{
\begin{tabular}{c|cccccccccc|cc}
\toprule
\multicolumn{13}{c}{{\textbf{Clear Weather $\rightarrow$ Rainy Weather}}} \\
\midrule
\multirow{3}*{Methods} & \multicolumn{10}{c|}{Average Precision$\uparrow$} &\multirow{3}*{ mAP$\uparrow$} & \multirow{3}*{NDS$\uparrow$}\\
\cmidrule{2-11}
&Car&Truck&{Cons. Vehicle}&Bus&Trailer&Barrier&Motorcycle&Bicycle&Pedestrian&{Cone}&&  \\

\midrule
PETR~\cite{liu2022petr} (Baseline) & 41.68 	&18.19 	&2.24 	&41.58 	&7.07 	&38.09 	&13.16 	&21.57 	&33.31 	&30.66 	&24.75 &32.84\\
SFA~\cite{wang2021exploring} &42.23 	&18.87 	&2.12 	&49.21 	&6.26 	&37.61 	&11.11 	&21.59 	&33.09 	&31.78 	&25.39  &33.18 \\
MTTrans~\cite{yu2022mttrans} &41.04 	&20.08 	&3.51 	&50.42 	&7.07 	&37.78 	&11.56 	&20.91 	&33.57 	&34.08 	&26.00 &33.68 \\
STM3D~\cite{li2022unsupervised} &42.45 	&19.22 	&2.68 	&47.62 	&7.00 	&38.52 	&13.39 	&21.44 	&33.85 	&31.97 	&25.81  &33.56 \\

DA-BEV (Ours) &46.72 	&25.06 	&3.38 	&51.62 	&8.70 	&46.65 	&17.46 	&26.85 	&39.37 	&37.80 	&30.36 &36.81\\
\end{tabular}
}
\resizebox{\linewidth}{!}{
\begin{tabular}{c|cccccccccc|cc}
\toprule
\multicolumn{13}{c}{{\textbf{Singapore $\rightarrow$ Boston}}} \\
\midrule
\multirow{3}*{Method} & \multicolumn{10}{c|}{Average Precision$\uparrow$} &\multirow{3}*{ mAP$\uparrow$} & \multirow{3}*{NDS$\uparrow$}\\
\cmidrule{2-11}
&Car&Truck&{Cons. Vehicle}&Bus&Trailer&Barrier&Motorcycle&Bicycle&Pedestrian&{Cone}&&  \\

\midrule
PETR~\cite{liu2022petr} (Baseline) &41.29 	&8.51 	&0.00 	&29.84 	&0.00 	&11.23 	&4.06 	&5.80 	&28.59 	&35.80 	&16.51 &21.74 \\
SFA~\cite{wang2021exploring} &42.75 	&10.51 	&0.00 	&29.25 	&0.00 	&12.78 	&3.68 	&8.43 	&31.70 	&37.04 	&17.61 &23.43\\
MTTrans~\cite{yu2022mttrans} &42.68 	&11.24 	&0.00 	&29.97 	&0.00 	&13.91 	&3.42 	&8.60 	&32.56 	&38.01 	&18.04 &24.69\\
STM3D~\cite{li2022unsupervised} &43.44 	&10.69 	&0.00 	&28.29 	&0.00 	&13.03 	&3.83 	&8.75 	&32.60 	&39.50 	&18.01 &24.68\\
DA-BEV (Ours) &45.35 	&14.69 	&0.00 	&31.68 	&0.00 	&19.58 	&6.63 	&9.78 	&34.20 	&42.12 	&20.40 	&28.35 \\
\bottomrule
\end{tabular}
}
\label{tab:MainResultDay2Night}
\end{table}

\begin{table}[tb]
\renewcommand{\arraystretch}{0.1}
\caption{
Unsupervised domain adaptation of camera-only BEV perception across different camera systems.
}
\tiny
\centering
\begin{tabular}{c|cccccc|cc}
\toprule
\multicolumn{9}{c}{{\textbf{Lyft $\rightarrow$ nuScenes Nighttime}}} \\
\midrule
\multirow{3}*{Methods} & \multicolumn{6}{c|}{Average Precision$\uparrow$} &\multirow{3}*{ mAP$\uparrow$} & \multirow{3}*{NDS*$\uparrow$}\\
\cmidrule{2-7}
&Car&Truck&Bus&Motorcycle&Bicycle&Pedestrian&  \\
\midrule
PETR~\cite{liu2022petr} (Baseline) &15.55&  1.00&  	1.83& 	8.53&  	0.43& 	4.26& 5.27 &10.54 \\
SFA~\cite{wang2021exploring} &16.50&  0.78&  1.45&  8.56& 0.44& 4.41& 5.35 &10.41\\
MTTrans~\cite{yu2022mttrans} &16.86&  1.62&  2.01&  8.11&  0.76& 2.87& 5.37 &10.20\\
STM3D~\cite{li2022unsupervised} &16.10&  1.06&  2.50& 10.48& 0.50& 4.06&  5.78 &11.06 \\
DA-BEV (Ours) &21.69& 1.57& 2.24&  13.79&  0.56& 7.75& 7.93 &13.62  \\
\toprule
\multicolumn{9}{c}{{\textbf{Lyft $\rightarrow$ nuScenes Rainy Weather }}} \\
\midrule
\multirow{3}*{Methods} & \multicolumn{6}{c|}{Average Precision$\uparrow$} &\multirow{3}*{ mAP$\uparrow$} & \multirow{3}*{NDS*$\uparrow$}\\
\cmidrule{2-7}
&Car&Truck&Bus&Motorcycle&Bicycle&Pedestrian&  \\
\midrule
PETR~\cite{liu2022petr} (Baseline) &22.91 &1.95 &2.56 &9.15 &0.88 &7.01 &7.41 &12.25 \\
SFA~\cite{wang2021exploring} &25.52&  1.92&  2.00&  9.95& 0.55& 9.88& 7.97& 13.19
 \\
MTTrans~\cite{yu2022mttrans} &24.84&  2.57&  2.08&  10.18&  1.63& 9.63& 8.48& 13.84
 \\
STM3D~\cite{li2022unsupervised} &26.28& 1.66& 1.92& 12.21& 1.51& 10.59 &9.03 & 14.41
  \\
DA-BEV (Ours) &30.46&  1.57&  2.74& 15.27&  1.53& 14.55& 11.02& 18.10 \\
\toprule
\multicolumn{9}{c}{{\textbf{Lyft $\rightarrow$ nuScenes}}} \\
\midrule
\multirow{3}*{Methods} & \multicolumn{6}{c|}{Average Precision$\uparrow$} &\multirow{3}*{ mAP$\uparrow$} & \multirow{3}*{NDS*$\uparrow$}\\
\cmidrule{2-7}
&Car&Truck&Bus&Motorcycle&Bicycle&Pedestrian&  \\
\midrule
PETR~\cite{liu2022petr} (Baseline) & 28.43 	&3.70 	&4.30 	&12.88 	&0.65 	&10.15 &10.02 &16.44\\
SFA~\cite{wang2021exploring} &33.17 	&3.55 	&4.18 	&12.44 	&0.62 	&11.96 	&10.98 &17.87
  \\
MTTrans~\cite{yu2022mttrans} &33.19 	&3.98 	&4.00 	&13.03 	&0.75 	&11.06 	&11.00 &17.98 \\
STM3D~\cite{li2022unsupervised} &36.51 	&4.34 	&4.63 	&15.96 	&0.51 	&14.93 	&12.81 &18.58  \\
DA-BEV (Ours) &38.45 	&3.93 	&4.23 	&23.53 	&0.75 	&22.85 	&15.62 &24.78\\
\bottomrule
\end{tabular}
\label{tab:MainLyft2nuScenes}
\end{table}

\subsection{Implementation Details}
We implement our DA-BEV upon PETR~\cite{liu2022petr} with backbone VoVNetV2~\cite{lee2020centermask}.
Following PETR, we adopt AdamW~\cite{loshchilov2017decoupled} optimizer with an initial learning rate of $1\times 10^{-4}$ and a weight decay of 0.01.
All models are trained for 50k iterations on one 2080Ti GPU with a batch size of 2.
For 3D object detection, the transformer decoder consist of 6 layers with 900 object queries.
For 3D scene segmentation, we follow the settings in \cite{zhou2022cross, liu2023petrv2}, where 625 segmentation queries along with a 6 layer transformer decoder are adopted for semantic segmentation result.

\begin{table}[!t]
\caption{
Ablation study of the proposed DA-BEV. 
The experiments are conducted on Daytime $\rightarrow$ Nighttime adaptation.
}
\renewcommand{\arraystretch}{0.15}
\centering
\tiny
\begin{tabular}{c|ccc|ccc|c|c}
\toprule
\multirow{3}{*}{Methods}&\multicolumn{3}{c|}{QAL} & \multicolumn{3}{c|}{QST}&\multirow{2}{*}{mAP$\uparrow$}&\multirow{2}{*}{NDS$\uparrow$} \\
\cmidrule{2-7}
 & Image-View & BEV & Co-Regularization & Image-View & BEV& Co-Regularization&&  \\

\midrule
PETR~\cite{liu2022petr} (Baseline) & -&-&-&-&-&-&15.93&22.41\\
\midrule
& \checkmark&&&&&&16.45 	&22.93 \\
& \checkmark&\checkmark&&&&&{16.96} 	&{23.21}   \\
& \checkmark&\checkmark&\checkmark&&&&{17.80}	&{24.07} \\
& \checkmark&\checkmark&\checkmark&\checkmark&&&18.53 &25.06 \\
& \checkmark&\checkmark&\checkmark&\checkmark&\checkmark&&19.39 	&26.30 \\
DA-BEV (Ours)& \checkmark&\checkmark&\checkmark&\checkmark&\checkmark&\checkmark&20.27&26.98 \\
\bottomrule
\end{tabular}
\label{tab:MainAblationStudy}
\end{table}

\subsection{Main Results}

Since there is little study on domain adaptive camera-only BEV, we compare DA-BEV with domain adaptation methods that achieved state-of-the-art performance in 2D object detection~\cite{wang2021exploring,yu2022mttrans} and monocular 3D object detection~\cite{li2022unsupervised}.

These compared methods can be easily applied on domain adaptive camera-only BEV perception by directly introducing their adversarial learning and/or self-training techniques into camera-only BEV perception model.
The comparisons are conducted over four challenging BEV adaptation scenarios as detailed in the following text.

Tables~\ref{tab:MainResultDay2Night} reports unsupervised domain adaptation of camera-only BEV perception across different illuminations, weathers and cities, respectively.
We can observe that DA-BEV achieves substantial improvements upon the baseline on these three adaptation scenarios, including $4.35\%$, $5.61\%$ and $3.89\%$ improvements in mAP, and $4.54\%$, $3.97\%$ and $6.61\%$ improvement in NDS.
In addition, DA-BEV outperforms the state-of-the-art UDA methods clearly and consistently across these three adaptation scenarios.
The superior BEV adaptation performance of DA-BEV is largely attributed to two factors: 1) DA-BEV introduces query-based domain adaptation that ingeniously exploits the complementary nature of image-view features and BEV features in BEV model, which helps tackle the domain gaps in BEV perception effectively.
2) DA-BEV introduces two query-based domain adaptation methods, i.e., QAL and QST, which complement each other and jointly achieve robust BEV perception adaptation effectively.

Table~\ref{tab:MainLyft2nuScenes} reports unsupervised domain adaptation of camera-only BEV perception across 
different camera systems.
Lyft and nuScences data are collected with very different camera systems that have clear domain gaps in terms of camera intrinsic parameters including camera focal lengths and field of views (FOVs) as well as camera extrinsic parameters such as camera poses and camera positions.
It can be observed that DA-BEV outperforms both the baseline on three adaptation scenarios substantially and consistently in mAP and NDC metrics, including over 2.66\%, 3.61\% and 5.60\% improvements in mAP and 3.08\%, 5.85\% and 8.34\% improvements in NDS.
These experiments demonstrate that DA-BEV can well handle the BEV adaptation across different camera systems that suffers from severe domain gaps from camera configuration inputs beyond those from image inputs.
On the other hand, it shows that our DA-BEV outperforms both baselines and other methods clearly on ``Lyft $\rightarrow$ nuScenes Nighttime" and ``Lyft $\rightarrow$ nuScenes Rainy Weather", demonstrating the effectiveness and robustness of DA-BEV under the cross-domain changes of camera configurations beyond image styles and appearance.


\subsection{Ablation Study}
We conduct extensive ablation studies to investigate  how different components in DA-BEV contribute. The experiments are conducted on Daytime $\rightarrow$ Nighttime adaptation.
As Table~\ref{tab:MainAblationStudy} shows, \textit{the baseline} does not perform well due to the domain gaps. 
As a comparison, including query-based adversarial learning on image-view features and/or BEV features improves the baseline clearly, indicating that QAL can simultaneously mitigate the inter-domain discrepancies in both 2D image-view features and 3D BEV features. 
In addition, introducing co-regularization between image-view and BEV query-based adversarial learning further improves the performance clearly, showing that image-view and BEV features encode complementary cross-domain discrepancy information that effectively regularize the adversarial learning of each other.

In addition, applying query-based self-training on image-view and/or BEV features brings clear improvements, indicating that QST helps learn effective 2D and 3D information of unlabeled target data from image-view and BEV, respectively.
Moreover, the co-regularization of query-based self-training of image-view and BEV features brings further improvement, demonstrating that image-view and BEV features provide complementary and comprehensive target information that effectively regularizes their self-training.
Further, the combination of QAL and QST performs the best clearly, showing that QAL and QST complement each other and jointly achieve robust unsupervised BEV perception adaptation effectively.

\subsection{Discussion}


\begin{table}[!t]
\caption{
Camera-only BEV perception of 3D scene segmentation on Daytime $\rightarrow$ Nighttime domain adaptation.
}
\renewcommand{\arraystretch}{0.5}
\tiny
\centering
\begin{tabular}{c|ccc|c}
\toprule
\multirow{3}{*}{Method} & \multicolumn{3}{c|}{IoU} & \multirow{3}{*}{mIoU} \\
\cmidrule{2-4}
                       &Drive	&Lane & Vehicle & \\

\midrule
PETR~\cite{liu2022petr} (Baseline)  &39.33 	&25.91 	&18.78 	&28.01\\
DA-BEV (Ours)   &51.37 	&30.43 	&24.98 	&35.59 \\
\bottomrule
\end{tabular}
\label{tab:BEVSemanticSeg}
\end{table}


\textbf{Generalzation across 3D perception tasks.}
We study how DA-BEV generalizes across 3D perception tasks by evaluating it over two representative tasks including 3D object detection and 3D scene segmentation. 
Experiments in Tables~\ref{tab:MainResultDay2Night} and ~\ref{tab:BEVSemanticSeg} show that DA-BEV achieves superior performance consistently over 3D detection and segmentation tasks, demonstrating the generalization ability of DA-BEV across different 3D perception tasks. 
\textbf{Parameter studies.} 
As defined in Eq.~\ref{eq:pseudolabelthresholdcal} parameter $\upsilon$ controls how many predictions we select as the pseudo labels.
We study $\upsilon$ by changing it from $10\%$ to $30\%$ with a step of $5\%$. The experiments in the top part of Table~\ref{tab:para_study} show that $\upsilon$ does not affect domain adaptation much while it changes from $15\%$ to $30\%$. 
The performance drops more when $\upsilon$ is set as $10\%$, largely because the adaptation is biased toward the source when very limited pseudo labels of target samples are selected.
As mentioned in Eq.~\ref{eq:ema_estimate_mu}, parameter $\gamma$ controls the update speed of the estimated global class probability distribution. We study $\gamma$ by changing it from 0.9 to 0.99999. The experiments in the bottom part of Table~\ref{tab:para_study} show that DA-BEV is tolerant to parameter $\gamma$ when it varies from 0.99 to 0.9999.
The performance drops more when $\gamma$ is set as 0.9 or 0.99999, largely because these two values lead to too fast or too slow update, resulting in unstable or nearly fixed estimation of global class probability distribution and degraded performance.

\textbf{Generalzation across network backbones.}
We study how DA-BEV generalizes across network backbones by evaluating it over three backbones of different sizes, including ResNet-50-C5 (46.9M), ResNet-50-P4 (46.9M) and VoV-P4 (96.9M).
Results in Table~\ref{tab:para_study} show that DA-BEV improves consistently over both small and large backbones, demonstrating the generalization ability of DA-BEV across different network backbones.

\begin{table}[!t]
\caption{
Left: Parameter analysis on Daytime $\rightarrow$ Nighttime adaptation. Right: Generalization across different backbones.}
\label{tab:para_study}
\renewcommand{\arraystretch}{0.5}
\tiny
\centering
\begin{tabular}{c|cc>{\columncolor{gray!25}}ccc}
\toprule
$\upsilon$ &10\% &15\% &{20}\% &25\% &30\%\\
\midrule
mAP   &18.24	&19.83	&20.27	&20.15	&19.57\\
\midrule
$\gamma$ &0.9 &0.99 & 0.999 & 0.9999 & 0.99999\\
\midrule
mAP &17.20 	&20.03 	&20.27 	&19.48 	&18.86\\
\bottomrule
\end{tabular}
\begin{tabular}{c|c|ccc}
\toprule
\multicolumn{2}{c|}{Backbone} &R50-C5 & R50-P4 &VoV-P4\\
\midrule
\multirow{1}*{PETR~\cite{liu2022petr} (Baseline)}  & mAP   &10.63 	&11.61   &15.93\\
\multirow{1}*{DA-BEV (Ours)}& mAP   &13.03 	&14.22   &20.27\\
\bottomrule
\end{tabular}
\end{table}

\begin{table}[!t]
\caption{
Generalization across different baselines.
}
\renewcommand{\arraystretch}{0.5}
\tiny
\centering
\resizebox{\linewidth}{!}{
\begin{tabular}{c|cccccccccc|cc}
\toprule
\multicolumn{13}{c}{{\textbf{Daytime $\rightarrow$ Nighttime}}} \\
\midrule
\multirow{3}*{Methods} & \multicolumn{10}{c|}{Average Precision$\uparrow$} &\multirow{3}*{ mAP$\uparrow$} & \multirow{3}*{NDS$\uparrow$}\\
\cmidrule{2-11}
&Car&Truck&{Cons. Vehicle}&Bus&Trailer&Barrier&Motorcycle&Bicycle&Pedestrian&{Cone}&&  \\

\midrule
PETR~\cite{liu2022petr} (Baseline)&40.79   &28.88   &0.41    &29.71   &0.00    &9.15  &22.00   &0.71    &24.66    &2.84    &15.92  &22.44 \\
DA-BEV (Ours) &45.07 	&41.96 	&1.91 	&38.12 	&0.00 	&16.66 	&25.74 	&0.54 	&28.05 	&4.65 	&20.27&26.98\\

\midrule
BEVFormer~\cite{yang2023bevformer} (Baseline)&39.31 &27.98 &0.18 &26.95 &0.00 &11.04, &21.84  &0.61& 26.50&  2.96& 15.73 &21.97
 \\
DA-BEV (Ours) &41.74&   36.41&   0.08&   36.38&   0.00&   19.47&    24.27&    0.40&    28.93&    3.79&  19.15& 24.14
\\
\midrule
CAPE~\cite{Xiong2023CAPE} (Baseline) 
&41.68& 27.95&  0.18& 29.57& 0.00& 11.49& 23.01&  0.98& 24.85&  3.20&  16.29 &23.54\\
DA-BEV (Ours) &44.81& 39.91&  0.19& 38.68&  0.00& 17.27& 25.98& 0.65& 28.09&  4.93& 20.05& 26.06\\
\bottomrule
\end{tabular}
}
\label{tab:AddtionalBaselineDay2Night}
\end{table}

\textbf{Generalization across different baselines.}
We study how our DA-BEV generalizes across different baselines by evaluating it over three types of mainstream BEV perception solutions, including PETR~\cite{liu2022petr}, BEVFormer~\cite{yang2023bevformer} and CAPE~\cite{Xiong2023CAPE}. 
For a fair comparison, all of them use VoVNetV2~\cite{lee2020centermask} as the backbone, and only single frame information is used for BEV perception. 
Experimental results in Table~\ref{tab:AddtionalBaselineDay2Night} show that DA-BEV brings clear improvements consistently over various baselines, demonstrating the generalization ability of DA-BEV across different BEV perception solutions. 

\textbf{Computation efficiency analysis.}
DA-BEV introduces an additional image-view decoder into the backbone of detector during training, 
and
this image-view decoder is discarded  in inference during testing. 
We study how this additional image-view decoder affects training and testing efficiency by 
benchmarking DA-BEV with several other methods 
in training time, testing time, training memory and testing memory.
Table \ref{tab: Computation efficiency analysis} reports 
the experiments on
``nuScences Daytime $\rightarrow$ nuScenes Nighttime'' adaptation over PETR~\cite{liu2022petr}.
It can be observed that incorporating an additional image-view decoder into PETR introduces 
slight
computation overhead during model training but has little 
effect on
model testing
, largely because
the image-view encoder is utilized during 
network
training 
but discarded
after adaptation.

\begin{table*}[t]
\caption{
Training and inference time analysis of all the compared methods. 
The experiments are conducted on one RTX 2080Ti.
}
\renewcommand{\arraystretch}{0.5}
\tiny
\centering
\resizebox{\linewidth}{!}{%
\begin{tabular}{c|cc|cc}
\toprule
Method     &Training Time (hours) &Training Memory (MB) &Testing Speed (images per second)  & Testing Memory (MB)\\
\midrule
PETR~\cite{liu2022petr} (Baseline)   &-    &-     &5.5551       &3075 \\
SFA~\cite{wang2021exploring}    &53.4966    &4967     &5.5551       &3077 \\
MMTrans~\cite{yu2022mttrans}    &56.3645    &10120    &5.5551       &3068\\
STM3D~\cite{li2022unsupervised} &53.6178    &4854     &5.5551       &3055\\
DA-BEV (Ours)                   &67.2670    &5540     &5.5551       &3089\\
\bottomrule
\end{tabular}
}
\label{tab: Computation efficiency analysis}
\end{table*}

\begin{figure}[t]
  \centering
  \includegraphics[width=0.9\linewidth]{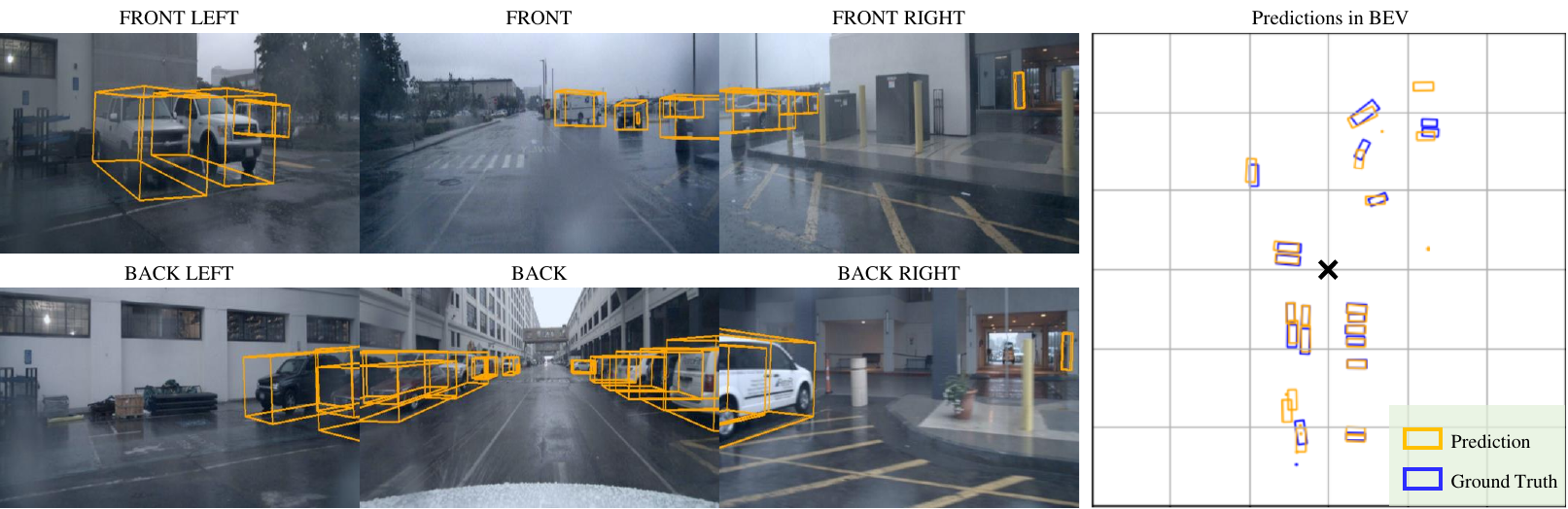}
  \caption{
 Qualitative illustration of DA-BEV on 3D object detection for cross-weather domain adaptation (i.e., Clear Weather $\rightarrow$ Rainy Weather).
  }
  \label{fig:short}
\end{figure}

\textbf{Qualitative results. }
Fig.~\ref{fig:short} provides qualitative illustrations which show that DA-BEV can produce accurate 3D object detection under weather-induced domain gaps, e.g., the leftmost car in Front Left camera becomes fuzzy due to rains and our DA-BEV can still detect it accurately.


\section{Conclusion}
In this paper, we present DA-BEV, the first domain adaptive camera-only BEV framework that addresses domain adaptive BEV challenges by exploiting the complementary nature of image-view features and BEV features.
DA-BEV introduces query-based adversarial learning (QAL) and query-based self-training (QST), where QAL/QST exploits the useful information queried from image-view features or BEV features to regularize the ``adversarial learning''/``self-training'' of the another.
Extensive experiments showcase the superior domain adaptive BEV perception performance of DA-BEV across various datasets and tasks.
Moving forward, we will further explore the complementary nature of image-view and BEV features by introducing the temporal information of them.

\section*{Acknowledgements}
This work was supported in part by the National Natural Science Foundation
of China under under Grant 62121001, Grant 62322117 and Grant 62371365; in part by Young Elite Scientist Sponsorship Program by the China Association for Science and Technology under Grant 2020QNRC001.

\bibliographystyle{splncs04}
\bibliography{main}
\end{document}